\documentclass[twocolumn,10pt]{asme2e}

\usepackage{subcaption}
\usepackage{epsfig} 
\usepackage{bm}
\usepackage{amssymb}
\usepackage{xcolor}
\usepackage{amsmath}
\usepackage{url}
\newcommand{\comment}[1]{}

\newtheorem{definition}{Definition} 
\confshortname{IDETC/CIE 2019}
\conffullname{the ASME 2019 International Design Engineering Technical Conferences \&\\
              Computers and Information in Engineering Conference}              

\confdate{August 18-21}
\confyear{2019}
\confcity{Anaheim, CA}
\confcountry{USA}

\papernum{DETC2019-97306}

\title{Towards Dynamic Simulation Guided Optimal Design of Tumbling Microrobots}

\author{Jiayin Xie
    \affiliation{
	Department of Mechanical Engineering\\
	Stony Brook University\\
	Stony Brook, New York 11794\\
    Email: jiayin.xie@stonybrook.edu
    }	
}

\author{Chenghao Bi
    \affiliation{
	School of Mechanical Engineering\\
    Purdue University\\
	West Lafayette, Indiana 47907\\
    Email: bi10@purdue.edu
    }
}

\author{David J. Cappelleri
    \affiliation{
	School of Mechanical Engineering\\
	Purdue University\\
	West Lafayette, Indiana 47907\\
    Email: dcappell@purdue.edu
    }
}

\author{Nilanjan Chakraborty \thanks{Address all correspondence to this author.}
    \affiliation{
	Department of Mechanical Engineering\\
	Stony Brook University\\
	Stony Brook, New York 11794\\
    Email: nilanjan.chakraborty@stonybrook.edu
    }
}

\begin{document}
\maketitle
\begin{abstract}
{\it Design of robots at the small scale is a trial-and-error based process, which is costly and time-consuming. There are no good dynamic simulation tools to predict the motion or performance of a microrobot as it moves against a substrate. At smaller length scales, the influence of adhesion and friction, which scales with surface area, becomes more pronounced. Thus, rigid body dynamic simulators, which implicitly assume that contact between two bodies can be modeled as point contact are not suitable. In this paper, we present techniques for simulating the motion of microrobots where there can be intermittent and non-point contact between the robot and the substrate. We use this simulator to study the motion of microrobots of different shapes and select shapes that are most promising for performing a given task.  }
\end{abstract}
\section*{INTRODUCTION}
Our overall long-term goal is to create a dynamic simulation tool that can be used to study the motion of microrobots of different geometry and manufacture a subset of the robots for experimentation. This will greatly help reduce the cost and effort of the microrobot design process, as the designer can hone in on the most promising designs.

\begin{figure}
\centering
\includegraphics[width=2in]{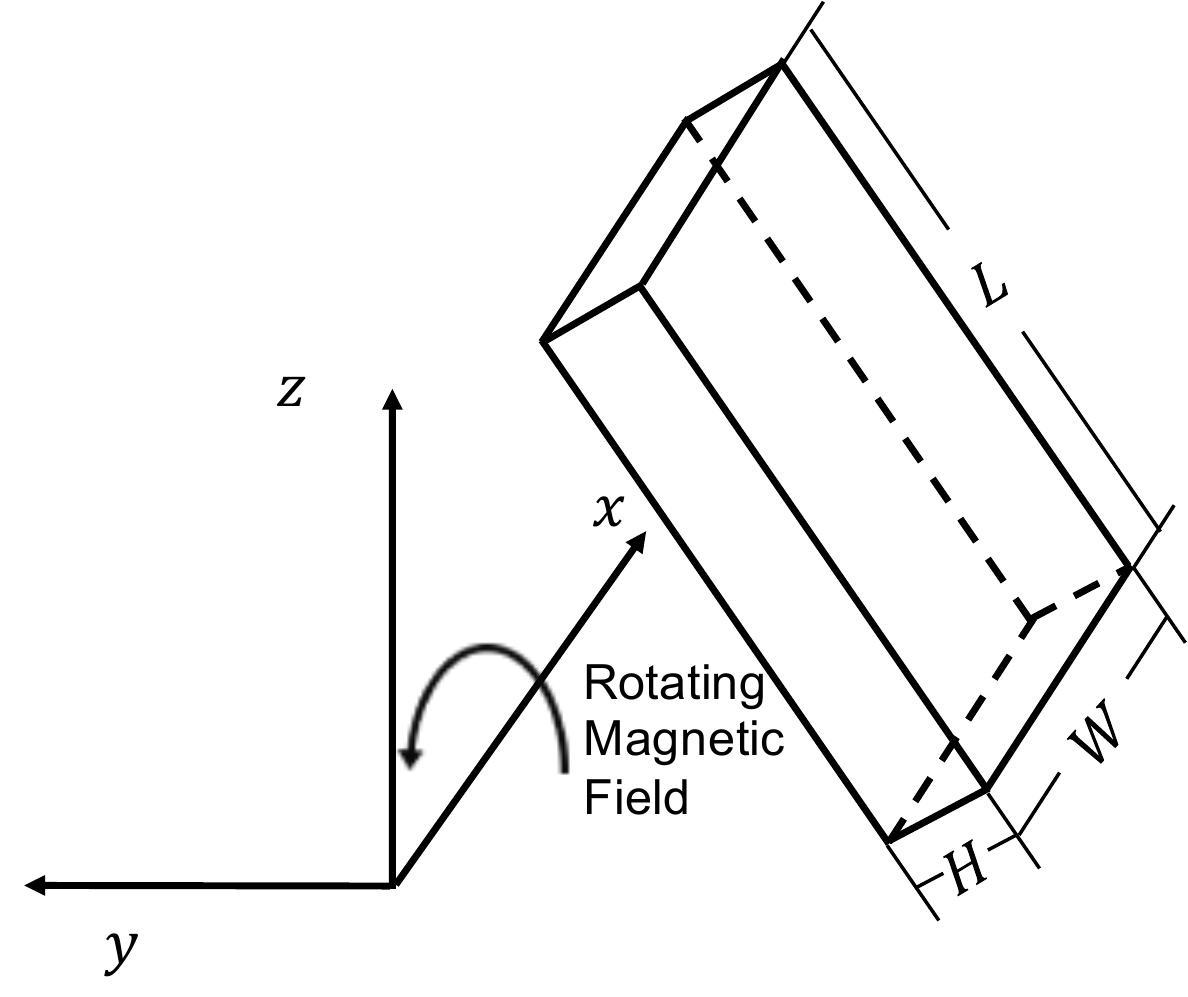}
\caption{Microscale magnetic tumbling ($\mu$TUM) robot tumbles on the planar surface. A magnetic field rotating about $x$ axis counterclockwisely, which causes the robot rotates in the same direction and tumbles forward (along the direction of $y$ axis). L, W and H represents the length, width and height of robot $\mu$TUM. } 
\label{figure_robot} 
\end{figure}

To validate the simulation tool at the microscale, our simulation results are compared against the performance of a mobile microrobot discussed in~\cite{Jing2013c},~\cite{Jing2013b}, and~\cite{Bi2018}, and shown schematically in Figure~\ref{figure_robot}. Differences in the orientation of the robot$'$s internal magnetization and that of a rotating external magnetic field induce a torque on the robot, causing it to tumble forward end-over-end. This tumbling locomotion has been shown to be versatile in both wet and dry environments on steep inclines and on rough surfaces~\cite{Bi2018}. It is especially promising for biomedical applications, due to the multiple complex environments within the human body that the robot can operate on. Tumbling microrobots have the potential to go to previously unreachable areas of the body and perform tasks such as targeted drug delivery, tissue biopsies, and toxin neutralization. Additionally, the external magnetic fields actuating the robot harmlessly penetrate living tissue and allow for tetherless locomotion.  One key limitation of external magnetic fields, however, is that they decrease volumetrically in strength as distance increases between the magnetic target and the source of the field. Therefore, it is beneficial to optimize the robot's design to achieve the most mobility under limited magnetic field strengths. It is also beneficial to optimize the design to travel over as many different surfaces as possible. A flexible simulation tool for virtual design iteration and optimization would be highly beneficial for this purpose. 

A critical challenge for simulating the tumbling microrobot is to model the intermittent and non-point contact between the robot and substrate, which will change during the motion based on the contact mode. For example, consider a curved shape microrobot tumbling over a planar surface. The contact mode between the robot and the surface is line contact, which changes as the robot moves. Most of the existing dynamic simulation methods~\cite{Pawashe2008,Pawashe2009a} implicitly assume that the contact between two bodies can be modeled as point contact. They choose contact points {\em a priori} in an ad hoc manner to represent the contact patch. For the curved robot, since the contact patch is time-varying it is not possible to choose contact point {a priori} and would thus introduce inaccuracy in simulation. Recently, we developed principled methods~\cite{xie2016rigid,Xie2018,xie2019}, to simulate contacting rigid bodies with planar convex and non-convex contact patches. In this paper, based on our previous work, we develop a method for simulating the motion of microrobots where the contact between the robot and the substrate is intermittment and non-point contact. Our contributions are as follows: (a) we extend our model in\cite{xie2016rigid} to handle the torque due to rotational magnetic field and surface area-dependent adhesive forces of a rigid body microrobot. (b) We present the procedure to compute the adhesive forces, which will change during motion based on the contact mode. (c) We also present numerical simulation results and perform preliminary comparisons with experimental results in several scenarios. Furthermore, we simulate the motion of microrobots with different shapes and choose shapes with best overall performances for given tasks.

\section*{RELATED WORK}
Past literature has demonstrated dynamic models for several mobile microrobots. Pawashe \textit{et al.} simulated a planar microrobot with stick-slip motion over dry horizontal surfaces \cite{Pawashe2008,Pawashe2009a}. The simulation was able to predict the robot's orientation and linear velocity over time under various external field parameters and surface properties. However, this model does not consider that the robot can tumble. Hu \textit{et al.} developed models for predicting the velocities of the rolling, walking, and crawling gaits of a soft-bodied magnetic millibot capable of multimodal locomotion~\cite{Hu2018}. The model helped determined which geometric dimensions were critical for the success of particular gaits of the robot. Morozov \textit{et al.} proposed a general theory to study the dynamics of arbitrarily-shaped magnetic propellers and rationalize previously unexplained experimental observations~\cite{Morozov2017}. To date, a comprehensive three-dimensional model that can predict a microrobot's trajectory and velocity over time with consideration of intermittent contact and inclined surfaces has yet to be developed.

In  this  paper,  we  present  techniques  for  simulating  motion of microrobots where there can be intermittent and non-point contact between the robot and the surface. The model we use is called a differential complementarity problem (DCP) model.
Let $\bm{u}\in \mathbb{R}^{n_1}$,  $\bm{v}\in \mathbb{R}^{n_2}$ and let $\bm{g}: \mathbb{R}^{n_1}\times \mathbb{R}^{n_2} \rightarrow \mathbb{R}^{n_1} $, $\bm{f}: \mathbb{R}^{n_1}\times \mathbb{R}^{n_2} \rightarrow \mathbb{R}^{n_2}$ be two vector functions and the notation $0 \le \bm{x} \perp \bm{y} \ge 0$ imply that $\bm{x}$ is orthogonal to $\bm{y}$ and each component of the vectors is non-negative. 
\begin{definition}
The differential (or dynamic) complementarity problem is to find $\bm{u}$ and $\bm{v}$ satisfying~\cite{Facchinei2007}:
$$\dot{\bm{u}} = \bm{g}(\bm{u},\bm{v}), \ \ \ 0\le \bm{v} \perp \bm{f}(\bm{u},\bm{v}) \ge 0 $$
\end{definition}
\begin{definition}
The mixed complementarity problem is to find $u$ and $v$ satisfying
$$\bm{g}(\bm{u},\bm{v})=0, \ \ \ 0\le \bm{v} \perp \bm{f}(\bm{u},\bm{v}) \ge 0$$
\end{definition}
If the functions $\bm{f}$ and $\bm{g}$ are linear, the problem is called a mixed linear complementarity problem (MLCP). Otherwise, the problem is called a mixed nonlinear complementarity problem (MNCP). As we will discuss later, our discrete-time dynamics model is a MNCP. 

Modeling the intermittent contact between bodies in motion as a complementarity constraint was first done by Lotstedt~\cite{Lotstedt82}. Subsequently, there was a substantial amount of effort in modeling and dynamic simulation with complementarity constraints~\cite{AnitescuCP96, Pang1996, StewartT96, DrumwrightS12, Todorov14}. The DCP that models the equations of motion usually can not be solved in closed form. Therefore, a time-stepping scheme has been introduced to solve the DCP. Depending on the assumptions made when forming the discrete equation of motions, the discrete-time model can be divided into a mixed linear complementarity problem (MNCP)~\cite{AnitescuP97, AnitescuP02} and a mixed non-linear complementarity problem~\cite{Tzitzouris01,chakraborty2014geometrically}. Furthermore, depending on whether the distance function between the two bodies (which is a nonlinear function of the configuration) is approximated or linearized, the time-stepping scheme can also be further divided into geometrically explicit schemes~\cite{AnitescuCP96, StewartT96} and geometrically implicit schemes~\cite{Tzitzouris01,chakraborty2014geometrically,xie2016rigid}. 

All of the time stepping schemes mentioned above assume the contact between two objects to be point contact. However, at the microscale, the influence of adhesion and friction become more pronounced. Both of these factors scale with the surface contact area. \comment{Therefore, the dynamic model, which takes non-point contact into account and can handle surface-area dependent adhesive force, is needed. }Recently, we presented a dynamic model that takes non-point contact (where the contact mode could be point contact, line contact, or surface contact) into account~\cite{xie2016rigid}. The model belongs to a geometrically implicit time-stepping scheme, in which the distance function depends on the geometry and configurations of the rigid body. In this paper, we extend this model to handle surface area-dependent adhesive forces of a rigid body microrobot that will change during motion based on the contact mode. The resulting discrete time model is a MNCP problem.

There has been much effort to model and understand the effect of non-point frictional contact~\cite{Erdmann1994, Goyal1991, Howe1996}. We use the so called soft-finger contact model~\cite{MLS94} for the general dynamic simulation. The model is based on a maximum power dissipation principle and it assumes all the possible contact forces or moments should lie within an ellipsoid. At the microscale, adhesion is more pronounced and can have a significant effect on microrobot locomotion. It is the combined effect of forces that may stem from capillary effects, electrostatic charging, covalent bonding, hydrogen bonding, Casimir forces, or Van der Waals interactions~\cite{526162}. All of these forces, aside from forces arising from electrostatic charging, become negligible outside of the nanometer range. Van der Waals forces, for example, primarily act at ranges of 0.2-20 nm ~\cite{Diller2011a}. These forces can also be unpredictable and difficult to model individually. Therefore, we clumped the forces together into a single adhesion force and assume its effect is insignificant if there is no direct contact between the microrobot and the substrate. We formulated this adhesive force as an empirical relationship where it is proportional to the surface contact area. This relationship is useful because our dynamic model is capable of predicting the time-varying surface contact area. Electrostatic force is treated as a constant, since the distance between the microrobot and the substrate undergoes minimal change as the robot moves.

\section*{DYNAMICS OF BODIES IN CONTACT}
In this section, we present an overview of the equations of motion of two rigid bodies in intermittent contact with each other. A microrobot moving on a surface may switch between having contact with the surface or no contact at all. Furthermore, when the robot is in contact, the contact may be a sliding or sticking contact (i.e., no relative velocity between the points on the objects in the contact region). Depending on the geometry of the robot and its configuration, the contact mode may also be point contact, line contact, or surface contact.  A key requirement for building dynamic simulators for the microrobots is the ability to handle surface area-dependent adhesive forces that will change during motion based on the contact mode. We will therefore use a complementarity-based model of dynamics that can handle the transition between no-contact and contact as well as sticking and sliding contact in a unified manner. Furthermore, since we can have non-point contact, we will use the equations of motion in~\cite{xie2016rigid} as our basic model for the dynamics.

\begin{figure}
\centering
\begin{subfigure}[b]{0.5\columnwidth}
\includegraphics[width=\columnwidth]{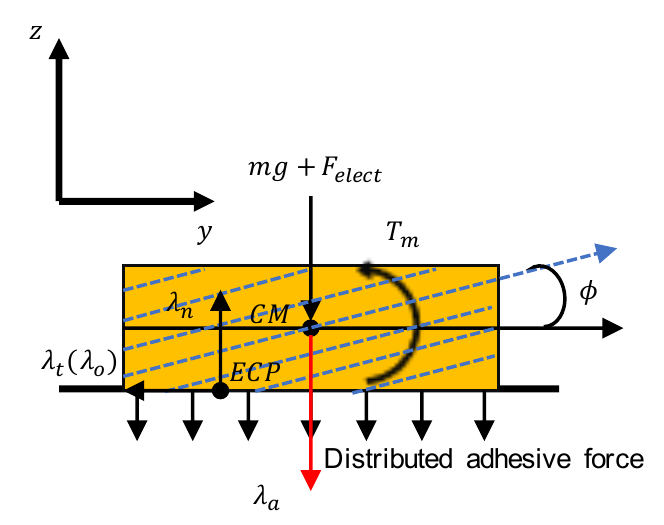}%
\caption{Robot in surface contact \\with horizontal surface in 2D.}
\label{figure_surface_contact} 
\end{subfigure}\hfill%
\begin{subfigure}[b]{0.5\columnwidth}
\includegraphics[width=\columnwidth]{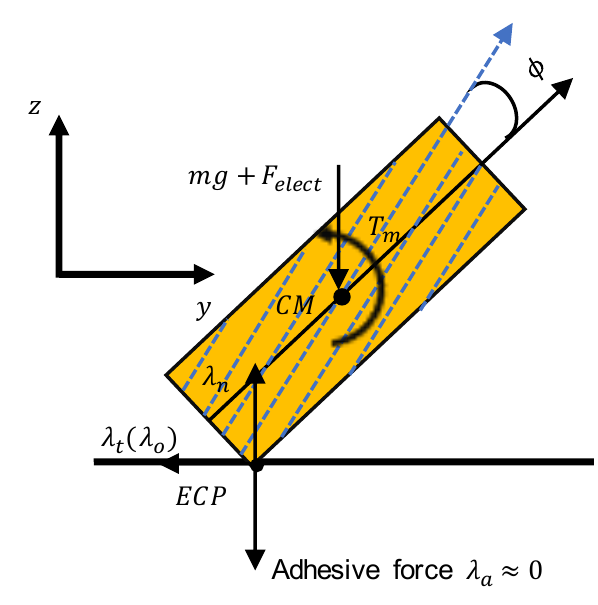}%
\caption{Robot in line contact with horizontal surface in 2D. }
\label{figure_line_contact} 
\end{subfigure}\hfill%
\caption{Force diagrams in 2D when robot has surface contact and line contact with the surface. The dashed lines in blue represent the internal magnetic alignment. The adhesive force is distributed uniformly over the surface area. When robot has line contact with the surface, the adhesive force is almost zero.}
\label{Contact_2D}
\end{figure}

The general equations of motion has three key parts: (i) Newton-Euler  differential equations of motion giving state update, (ii) algebraic and complementarity constraints modeling the fact that two rigid bodies cannot penetrate each other, and (iii) model of the contact force and moments acting on the contact patch. For general rigid body motion, the model of contact forces and moments use Coulomb's assumption that {\em the normal force acting between two objects is independent of the nominal contact area between the two objects}. This is a reasonable assumption for nominally rigid objects at macroscopic length scales, where the inertial forces are dominating. However, at the length-scale of microrobots, the force of adhesion between the contacting surfaces is comparable to inertial forces. So, the contact model should also take into consideration the effect of the surface-area dependent forces. These forces, combined under a single adhesive force, are illustrated in Figure~\ref{Contact_2D}.

For simplicity of exposition, we assume one body to be static. 
Let $\bm{V} = [\bm{v}^T ~\bm{\omega}^T]^T$ be the generalized velocity of the rigid body, where $\bm{v} \in \mathbb{R}^3$ is the linear velocity and $\bm{\omega} \in \mathbb{R}^3$ is the angular velocity of the rigid body. Let $\bm{q}$ be the configuration of the rigid body, which is a concatenated vector of the position and a parameterization of the orientation of the rigid body. 

{\bf Newton-Euler Equations of Motion}:
The Newton-Euler equations of motion of the rigid body are:
\begin{equation} \label{eq_general}
\bm{M}(\bm{q})
{\dot{\bm{V}}} = 
\bm{W}_{n}\lambda_{n}+
\bm{W}_{t}\lambda_{t}+
\bm{W}_{o}\lambda_{o}+
\bm {W}_{r}\lambda_{r}+
\bm{\lambda}_{app}+\bm{\lambda}_{vp}
\end{equation}
where $\bm{M}(\bm{q})$ is the inertia tensor, $\bm{\lambda}_{app}$ is the vector of external forces and moments (including gravity), $\bm{\lambda}_{vp}$ is the  centripetal and Coriolis forces. The magnitude of the normal contact force is $\lambda_n$. The magnitude of tangential contact forces are $\lambda_t$ and $\lambda_o$. The magnitude of the moment due to the tangential contact forces about the contact normal is $\lambda_r$. The vectors $\bm{W}_n$, $\bm{W}_t$, $\bm{W}_o$ and $\bm{W}_r$ map the contact forces and moments from the contact point to the center of mass of the robot. The expressions of $\bm{W}_n$, $\bm{W}_t$, $\bm{W}_o$ and $\bm{W}_r$ are:
\begin{equation}
\begin{aligned}
\label{equation:wrenches}
\bm{W}_{n} =  \left [ \begin{matrix} 
\bm{n}\\
\bm{r}\times \bm{n}
\end{matrix}\right],
\bm{W}_{t} =  \left [ \begin{matrix} 
\bm{t}\\
\bm{r}\times \bm{t}
\end{matrix}\right],
\bm{W}_{o} =  \left [ \begin{matrix} 
\bm{o}\\
\bm{r}\times \bm{o}
\end{matrix}\right],
\bm{W}_{r} =  \left [ \begin{matrix} 
\bm{0}\\
\ \ \bm{n} \ \
\end{matrix}\right]
\end{aligned}
\end{equation}
where $(\bm{n},\bm{t},\bm{o}) \in \mathbb{R}^3$ are the axes of the contact frame, $\bm{0} \in \mathbb{R}^3$ is a column vector with each entry equal to zero. As shown in Figure~\ref{Contact_2D}, vector $\bm{r} = [a_x-q_x,a_y-q_y,a_z-q_z]$ is the vector from equivalent contact point (ECP) $\bm{a}$, to center of mass (CM), where $(q_x, q_y, q_z)$ is the position of the CM. In the next section, we will provide definition for the equivalent contact point (ECP). Please note that Equation~\eqref{eq_general} is a system of $6$ differential equations.

{\bf Modeling Rigid Body Contact Constraints:} The contact model that we use is a complementarity-based contact model as described in~\cite{xie2016rigid,chakraborty2014geometrically}. In~\cite{xie2016rigid}, we introduced the notion of an equivalent contact point (ECP) to model non-point contact between objects. 
\begin{definition}
 Equivalent Contact Point (ECP) is a unique point on the contact surface that can be used to model the surface (line) contact as point contact  where  the  integral of the total moment (about the point) due to the distributed normal force on the contact patch is zero.
\end{definition}
 The ECP defined here is the same as the center of friction. Now let's describe the contact model mathematically. Let two objects $F$ and $G$ be defined by intersection of convex inequalities $f_{i}(\bm{\zeta}_1)\le 0, i = 1,..,m$, and $g_{j}(\bm{\zeta}_2)\le 0, j = m+1,..,n$ respectively. Let $\bm{a}_1$ and $\bm{a}_2$ be pair of ECP's or closest points (when objects are separate) on F and G, respectively. The complementarity conditions for nonpenetration can be written as either one of the following two sets of conditions~\cite{chakraborty2014geometrically}:
\begin{equation}
\begin{aligned}
\label{equation:contact_multiple_comp}
0 \le \lambda_{n} \perp \mathop{max}_{1,...,m} f_{i}(\bm{a}_2) \ge 0\\
0 \le \lambda_{n} \perp \mathop{max}_{j=m+1,...,n}g_{j}(\bm{a}_1) \ge 0
\end{aligned}
\end{equation}

The solution of ECP's $\bm{a}_1$ and $\bm{a}_2$ is given by the following minimization problem:
\begin{equation}
\label{equation:optimazation}
(\bm{a}_1,\bm{a}_2) = arg \min_{\bm{\zeta}_1,\bm{\zeta}_2}\{ \|\bm{\zeta}_1-\bm{\zeta}_2 \| \ f_{i}(\bm{\zeta}_1) \le 0,\ g_{j}(\bm{\zeta}_2) \le 0 \}
\end{equation}
where $i =1,...,m$ and $j=m+1,...,n$.

Using a slight modification of the KKT conditions for the optimization problem in Equation~\eqref{equation:optimazation}, and combing it with either one of the conditions in Equation~\eqref{equation:contact_multiple_comp}, we get the complete contact model between two rigid bodies:
\begin{equation}
\begin{aligned}
\label{equation:re_contact_multiple1}
&\bm{a}_{1}-\bm{a}_{2} = -l_{k}\mathcal{C}(\bm{F},\bm{a}_1), \
\mathcal{C}(\bm{F},\bm{a}_1)= -\mathcal{C}(\bm{G},\bm{a}_2)\\
0 \le &\left[ \begin{matrix} l_{i}\\ l_{j}\\ \lambda_{n}  \end{matrix} \right] \perp \left[ \begin{matrix} &-f_{i}(\bm{a}_{1}), \quad i = 1,...,m \\ &-g_{j}(\bm{a}_{2}), \quad j = m+1,...,n\\ &\max\limits_{i=1,...,m} f_{j}(\bm{a}_2)  \end{matrix} \right] \ge 0 
\end{aligned}
\end{equation}
where $k$ is the index of active constraint on body $F$, and the normal cones are: $\mathcal{C}(\bm{F},\bm{a}_1) = \nabla f_{k}(\bm{a}_{1})+\sum_{i = 1,i\neq k}^m l_{i}\nabla f_{i}(\bm{a}_{1})$, $\mathcal{C}(\bm{G},\bm{a}_2) = \sum_{j = m+1}^n l_{j} \nabla g_{j} (\bm{a}_{2})$.

\begin{figure}
\centering
\includegraphics[width=2in]{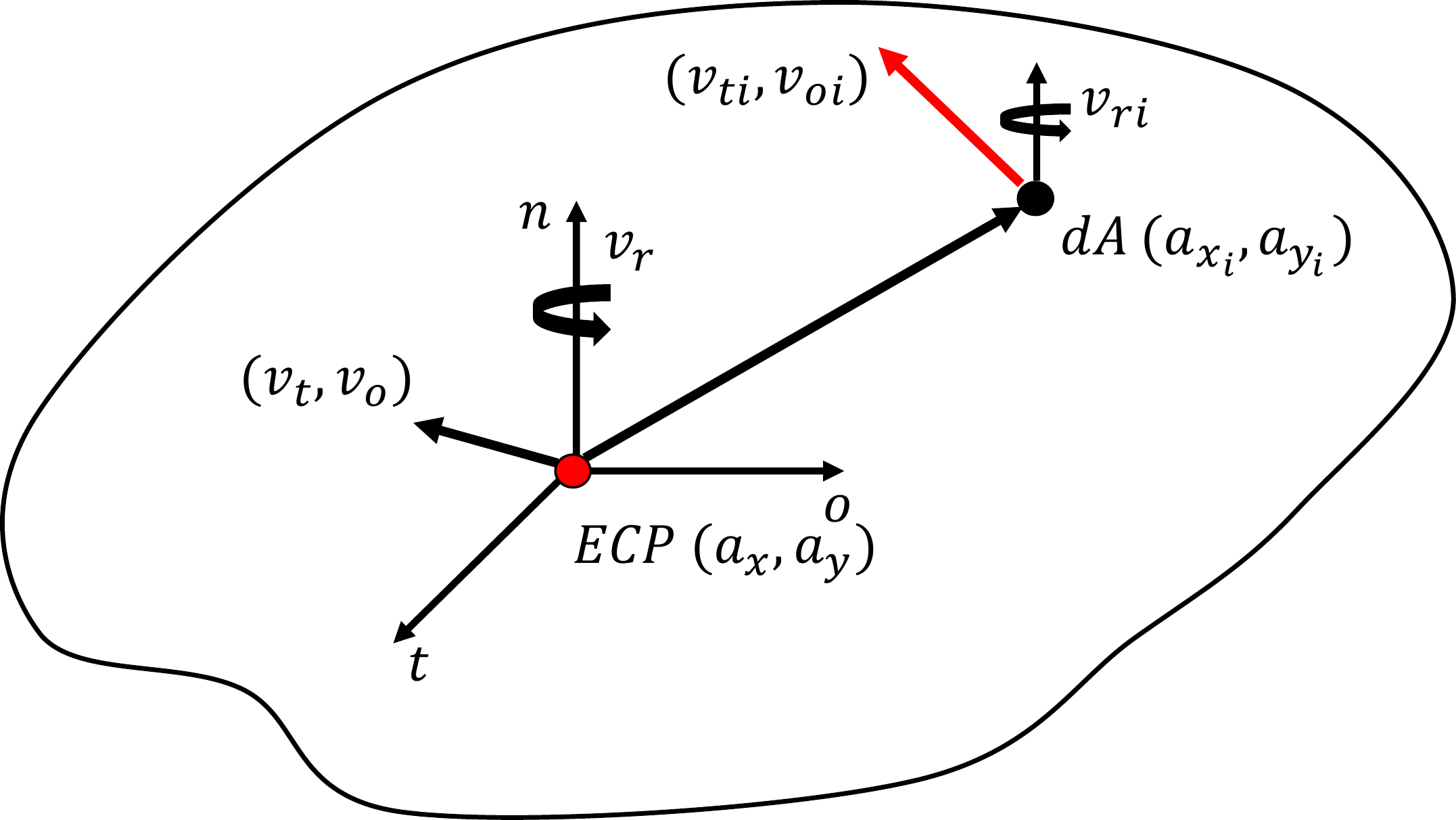}
\caption{Notation for planar sliding motion. } 
\label{figure_pressure} 
\end{figure} 

{\bf Friction Model:} We use a friction model based on the maximum power dissipation principle, which has been previously proposed in the literature for point contact~\cite{Moreau1988}. The maximum power dissipation principle states that among all the possible contact forces and moments that lie within the friction ellipsoid, the forces that maximize the power dissipation at the contact are selected. For non-point contact, we will use a generalization of the maximum power dissipation principle, where, we select contact forces/moments and contact velocities that maximize the power dissipation over the entire contact patch. We will now show that the problem formulation using the power loss over the whole contact patch can be reduced to the friction model for point contact with the ECP as the chosen point. Mathematically, the power dissipated over the entire surface, $P_c$ is 
\begin{equation}
\label{eq:power}
P_c = -\int_{A} (v_{ti}\beta_{ti}+v_{oi}\beta_{oi}+v_{ri}\beta_{ri})dA
\end{equation}
where $v_{ti},v_{oi},v_{ri}$ are the sliding velocity at $dA$, $\beta_{ti},\beta_{oi}$ are the frictional force per unit area and $\beta_{ri}$ is the resistive moment per unit area at $dA$, about the normal to the contact patch. We will assume a planar contact patch which implies that the contact normal is the same at all points on the contact patch. As shown in Figure~\ref{figure_pressure}, the angular velocity is constant across the patch, i.e., $v_{ri} = v_r$, for all $i$. Let $v_t$ and $v_o$ be the components of tangential velocities at the ECP. From basic kinematics, we know that $v_{ti} = v_t - v_{ri}(a_{yi}-a_y)$ and $v_{oi} = v_o + v_{ri}(a_{xi}-a_x)$, where ($a_x$, $a_y$) are the $x$ and $y$ coordinates of the ECP and ($a_{xi}$, $a_{yi}$) are the $x$ and $y$ coordinates of a point on the patch. Substituting the above in Equation~\eqref{eq:power} and simplifying, we obtain
\begin{equation}
\label{eq:power2}
P_c = -\left[\int_{A} v_{t}\beta_{ti}dA + \int_{A}v_{o}\beta_{oi}dA +\int_{A}v_{ri}\beta_{ri}^{\prime}dA \right ]
\end{equation}
where $\beta^{\prime}_{ri} = \beta_{ri} -\beta_{ti}(a_{yi}-a_y) + \beta_{oi}(a_{xi}-a_x)$.
%
%
By noting that $\int \beta_{ti}dA = \lambda_t, \int \beta_{oi}dA  = \lambda_o, \int \beta^{\prime}_{ri}dA  = \lambda_r$, where $\lambda_t$, $\lambda_o$ are the net tangential forces at the ECP and $\lambda_r$ is the net moment about the axis normal to the contact patch and passing through the ECP, the power dissipation over the entire contact patch is given by $P_c = - (v_t \lambda_t + v_o\lambda_o + v_r \lambda_r)$. For specifying a friction model, we also need a law or relationship that bounds the magnitude of the friction forces and moments in terms of the magnitude of the normal force~\cite{goyal1991planar}. Here, we use an ellipsoidal model for bounding the magnitude of tangential friction force and friction moment. This friction model has been previously proposed in the literature~\cite{goyal1991planar,Moreau1988,xie2016rigid,chakraborty2014geometrically} and has some experimental justification~\cite{howe1996practical}.
Thus, the contact wrench is the solution of the following optimization problem: 
\begin{equation}
\begin{aligned}
\label{equation:friction}
{\rm max} \quad -(v_t \lambda_t + v_o\lambda_o + v_r \lambda_r)\\
{\rm s.t.} \quad \left(\frac{\lambda_t}{e_t}\right)^2 + \left(\frac{\lambda_o}{e_o}\right)^2+\left(\frac{\lambda_r}{e_r}\right)^2 - \mu^2 \lambda_n^2 \le 0
\end{aligned}
\end{equation}
where the magnitude of contact force and moment at the ECP, namely, $\lambda_t$, $\lambda_o$, and $\lambda_r$ are the optimization variables. The parameters, $e_t$, $e_o$, and $e_r$ are positive constants defining the friction ellipsoid and $\mu$ is the coefficient of friction at the contact~\cite{howe1996practical,trinkle1997dynamic}.
As stated before, we use the contact wrench at the ECP to model the effect of entire distributed contact patch.\comment{ Therefore $v_t$ and $v_o$ are the tangential components of velocity at the ECP; $v_r$ is the relative angular velocity about the normal at ECP. } 
Note that there is {\em no assumption made on the nature of the pressure distribution between the two surfaces}. A key aspect of this work which is different from previous effort that is here we consider that the normal force can be a function of the contact surface area. We will elaborate on how this is done within the context of the discrete-time framework, since this requires that we identify the contact surface as part of our dynamic simulation algorithm.  

Using the Fritz-John optimality conditions of Equation~\eqref{equation:friction}, we can write~\cite{trinkle2001dynamic}:
\begin{equation}
\begin{aligned}
\label{eq:friction}
0&=
e^{2}_{t}\mu \lambda_{n} 
v_t+
\lambda_{t}\sigma \\
0&=
e^{2}_{o}\mu \lambda_{n}  
v_o+\lambda_{o}\sigma \\
0&=
e^{2}_{r}\mu \lambda_{n}v_r+\lambda_{r}\sigma\\
0& \le \mu^2\lambda_n^2- \lambda_{t}^2/e^{2}_{t}- \lambda_{o}^2/e^{2}_{o}- \lambda_{r}^2/e^{2}_{r} \perp \sigma \ge 0
\end{aligned}
\end{equation}
where $\sigma$ is a Lagrange multiplier corresponding to the inequality constraint in~\eqref{equation:friction}. 
\section*{EQUATIONS OF MOTION FOR TUMBLING MICROROBOT}

As shown in Figure~\ref{figure_robot}, the magnetic microscale tumbling robot ($\mu$TUM) presented in this paper is cuboid-shaped and embedded with magnetic particles. The robot's magnetic features are aligned along a certain direction and optimally it should to be aligned along lengthwise direction of the robot. An alignment offset angle is defined when there exists an angular difference between actual alignment direction and the desired alignment direction.

There exists one magnetic field which rotates counterclockwise about the $x$ axis of the world frame. When the magnetic alignment of the field differs from that of the robot, a magnetic torque is applied on the robot until it is realigned with the field. Therefore, a rotating magnetic field causes the robot to rotate about the same axis. As shown in Figure~\ref{figure_robot}, if the robot is resting on the surface, the rotating field causes the tumbling motion of the robot, i.e., the robot will move forward by continuously flipping end-over-end.

{\bf Notation}:
The following notation will be used for defining the problem mathematically:

\begin{itemize}
\item[$\circ$] $L, M, H$ $-$  length, width and height of the robot
\item[$\circ$] $\bm{M} = diag(m,m,m,I_{xx},I_{yy},I_{zz})$ $-$ inertia tensor of the robot, where $m$ represents mass and $I_{(.)}$ represents moment of inertia in body frame
\item[$\circ$] $F_{elect}$ $-$ electrostatic force of the robot
\item[$\circ$] $V_m$ $-$ magnetic volume of the robot
\item[$\circ$] $\bm{E}\in \mathbb{R}^3$ $-$ magnetization of the robot (The blue dashed lines in Figure~\ref{Contact_2D})
\item[$\circ$] $\phi$ $-$ magnetic alignment offset angle
\item[$\circ$] $\bm{B}\in \mathbb{R}^3$, $\bm{T}_m \in \mathbb{R}^3$ $-$ magnetic field strength and torque
\item[$\circ$] $f_{rot}$ $-$ the frequency of rotational field 
\item[$\circ$] $\mu$ $-$ friction coefficient between robot and surface
\item[$\circ$] $C$ $-$  coefficient of adhesive force between robot and surface
\item[$\circ$] $A_{contact}$ $-$ area of contact region between robot and surface
\item[$\circ$] $\lambda_{a}$ $-$ the adhesive force between robot and surface
\item[$\circ$] $e_t, e_o, e_r$ $-$ friction parameters defining the friction ellipsoid
\item[$\circ$] $\bm{n}\in \mathbb{R}^3$ $-$ the contact normal, which is used to define the normal axis of contact frame
\item[$\circ$] $\bm{t}\in \mathbb{R}^3$, $\bm{o}\in \mathbb{R}^3$ $-$ tangential axies of the contact frame
\item [$\circ$] $\bm{v} = [v_x, v_y, v_z]$ $-$ linear velocity of the robot
\item [$\circ$] $\bm{w} = [w_x, w_y, w_z]$ $-$ angular velocity of the robot
\item [$\circ$] $\lambda_n, \lambda_t, \lambda_o$  $-$ normal and tangential contact forces
\item [$\circ$] $\lambda_r$ $-$ frictional moment about contact normal $\bm{n}$
\item [$\circ$] $\bm{a}_1 \in \mathbb{R}^3, \bm{a}_2 \in \mathbb{R}^3$ $-$ pair of equivalent contact points (ECP)
\item [$\circ$] $\sigma$ $-$ Lagrange multiplier associated with the friction model, which represents the magnitude of slip velocity
\item [$\circ$] $\bm{l_1} =[l_1,...,l_m], \bm{l_2} =[l_{m+1},...,l_n]$ $-$ Lagrange multipliers in contact constraints
\end{itemize}

The magnetic torque $\bm{T}_m$ applied to the microrobot is:
\begin{equation}
\label{Torque}
    \bm{T}_m = V_m\bm{E}\times\bm{B}
\end{equation}

 The direction of $\lambda_{a}$ is along negative direction of $\bm{n}$, and its value depends on the material of the object and the area of contact region. The expression for $\lambda_{a}$ is:
\begin{equation}
\label{ADH}
\lambda_{a} = CA_{contact}  
\end{equation}

{\bf Newton-Euler Equations for Tumbling Microrobot}: 
As shown in Figure~\ref{Contact_2D}, the generalized applied force $\bm{\lambda}_{app} \in \mathbb{R}^6$ acting on CM of the robot includes gravity force $mg$, electrostatic force $F_{elect}$, adhesive force $\lambda_a$ and magnetic torque $\bm{T}_m \in \mathbb{R}^3$. The contact wrench acting on the ECP includes normal contact force, $\lambda_n$, and frictional forces and moments, $\lambda_t, \lambda_o$ and $\lambda_r$. The generalized velocity is $\bm{V} = [\bm{v},\bm{w}]$. The Newton-Euler equations are:
\begin{equation}
\label{NE}
\bm{M} \dot{\bm{\nu}} = \bm{W}
\left[\begin{matrix}
\lambda_{n}\\
\lambda_{t}\\
\lambda_{o}\\
\lambda_{r}
\end{matrix}
\right]
+
\left [\begin{matrix}
0\\
0\\
-(mg+F_{elect}+\lambda_a)\\
\bm{T}_m
\end{matrix} \right]
+\bm{\lambda}_{vp}
\end{equation}
where the mapping matrix $\bm{W} = [\bm{W}_{n}, \bm{W}_{t}, \bm{W}_{o}, \bm {W}_{r}] \in \mathbb{R}^{6\times 4}$ is computable based on Equation~\eqref{equation:wrenches}. The magnetic torque $\bm{T}_m$ is based on Equation~\eqref{Torque}. Please note that Equation~\eqref{NE} is a system of 6 differential equations.

{\bf Discrete-time dynamic model}:We use a velocity-level formulation and an Euler time-stepping scheme to discretize the above system of equations. Let superscripts $u$ be the beginning of current time step, $u+1$ be the end of current time step, and $h$ be the time step length. Let $\dot{\bm{V}}  \approx (\bm{V}^{u+1} -\bm{V}^u)/h$ and impulse $p_{(.)} = h\lambda_{(.)}$, we get the following discrete-time system. The system of equations in general is a mixed nonlinear complementarity problem. The vector of unknowns, $\bm{z}$, can be partitioned into $\bm{z} = [\bm{u}_z,\bm{v}_z]$, where:
\begin{equation*}
    \bm{u}_z = [\bm{V};\bm{a}_1;\bm{a}_2;p_t;p_o;p_r], \ \bm{v}_z = [\bm{l}_1;\bm{l}_2;\sigma;p_n]
\end{equation*}

The equality constraints in the mixed NCP are:
\begin{equation}
\begin{aligned}
&\bm{M}^u (\bm{V}^{u+1}- \bm{V}^{u}) =  \bm{W}^{u+1}
\left[\begin{matrix}
p^{u+1}_{n}\\
p^{u+1}_{t}\\
p^{u+1}_{o}\\
p^{u+1}_{r}
\end{matrix}
\right]
-
\left [\begin{matrix}
0\\
0\\
mgh+p_{elect}+p^{u}_a\\
-\bm{T}^u_mh
\end{matrix} \right]
 - \bm{p}_{vp}^{u} \\
&0 = \bm{a}^{u+1}_1-\bm{a}^{ u+1}_2+l^{u+1}_{k} \mathcal{C}(\bm{F},\bm{a}^{u+1}_{1})\\
&0 = \mathcal{C}(\bm{F},\bm{a}^{u+1}_{1})+\mathcal{C}(\bm{G},\bm{a}^{  u+1}_{2})\\
&0 = \mu e_t^2p^{u+1}_{n}{\bm{W}^{T u+1}_{t}}\bm{V}^{u+1}+p^{u+1}_{t}\sigma^{u+1}\\
&0 = \mu e_o^2p^{u+1}_{n}\bm{W}^{T u+1}_{o}\bm{V}^{u+1}+p^{u+1}_{o}\sigma^{u+1}\\
&0 = \mu e_r^2p^{u+1}_{n}\bm{W}^{T u+1}_{r}\bm{V}^{u+1}+p^{u+1}_{r}\sigma^{u+1}
\end{aligned}
\end{equation}
The complementarity constraints on $\bm{v}_z$ are:
\begin{equation}
\begin{aligned}
0 \le &\left[ \begin{matrix} \bm{l}^{u+1}_{1}\\ \bm{l}^{u+1}_{2}\\  \sigma^{u+1} \\p^{u+1}_{n}  \end{matrix} \right] \perp \left[ \begin{matrix} -\bm{f}(\bm{a}^{u+1}_{1}) \\-\bm{g}(\bm{a}^{ ^{u+1}}_{2})\\  \xi  \\ \max \bm{f}(\bm{a}^{ ^{u+1}}_2) \\
\end{matrix} \right] \ge 0 
\end{aligned}
\end{equation}
where $\xi =(\mu p^{u+1}_{n})^2-(p^{u+1}_{t}/e_t)^2 -(p^{u+1}_{o}/e_o)^2 -(p^{u+1}_{r}/e_r)^2$. Furthermore, the adhesive impulse $p^u_{a}$ is required as input at the beginning of each time step. We can compute $p^u_{a}$ based on Equation~\eqref{ADH}. However, in order to compute $p^u_{a}$, we need to know the contact are at each time step. However, this is not part of our solution to the dynamic model. In next section, we will discuss the procedure to compute the contact area, $A_{contact}$.

{\bf Computing the area of contact region}:
In general, the area of contact region, $A_{contact}$, depends on the geometry and configurations of objects in contact, which is hard to describe mathematically. However, in our case, the contact happens between the microrobot ($\mu$TUM) and the planar surface. The contact region is the side of the robot in contact with the surface. The geometry and dimension of the robot can be measured a priori and we can compute the area of each side of the robot. The next question is: which side of the robot is in contact at the current time?

The question can be answered by utilizing Lagrange multipliers of contact constraints. Based on the complementary condition, once  $l^{u+1}_i > 0$,  its associated constraint $f_{i}(\bm{a}^{u+1}_{1}) =0$, i,e., the Equivalent contact point should lie on the constraint or side $i$. If $p^{u+1}_n > 0$, which indicates robot has contact on the surface at the end of the current time, the active constraint or side $i$ will be the side of robot that has contact with the surface. 

To sum up, first we can compute the area of each side of the robot based on the knowledge of robot's geometry and dimensions. Then, we solve the discrete-time model at each time step. The solutions for $l^{u+1}_i$ and $p_n^{u+1}$ will be utilized to identify the side or boundary of the robot on contact and return us $A^{u+1}_{contact}$. Eventually, based on Equation~\eqref{ADH}, we compute adhesive impulse $p^{u+1}_a$, which would be used as input for next time step.

\vspace{-0.10in}
\section*{NUMERICAL RESULTS}

\begin{figure}[h]
\centering
\includegraphics[width=3in]{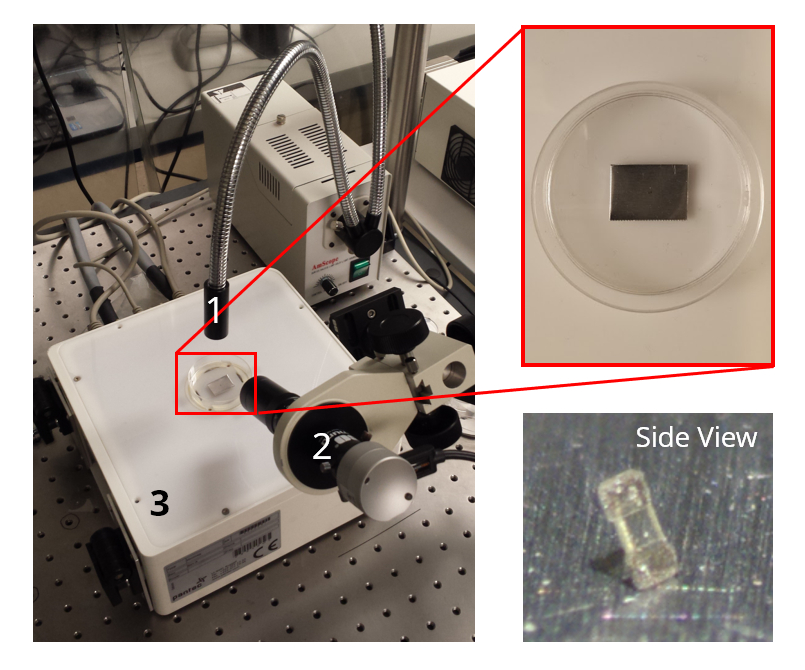}
\caption{Experimental setup with halogen lamp (1), side camera (2) and MFG-100 system (3). Additional images show an aluminum surface inside a petri dish at the center of the workspace and a side view of the {$\mu$}TUM as seen through the camera.}
\label{figure:Experimental_Setup} 
\vspace{-0.25in}
\end{figure}

{\bf Experimental Setup}: To validate our dynamic model, we compared experimental results against our simulated results. The experiment microrobots are composed of two SU-8 polymer ends doped with magnetic NdFeB particles and a non-magnetic middle section that is entirely made up of SU-8 polymer. Their external dimensions are: Length $L = 0.8\times 10^{-3}m$, Width $W =0.4\times 10^{-3}m$, and Height $H = 0.1\times 10^{-3}m$. The material properties are listed in Table~\ref{table_1} and they were fabricated using a two-step photolithography process described in \cite{Bi2018}. Additional robots were fabricated with the parameters listed in Table~\ref{table_3}. These robots underwent an additional step where they were exposed to a 9 T uniform magnetic field generated by a PPMS machine (Quantum Design) after the SU-8 curing process. This field was strong enough to realign the embedded NdFeB particles homogeneously and the resulting magnetization was measured using the same machine. A system of eight electromagnetic coils (MFG-100 system, MagnetibotiX AG) was used to generate the rotating magnetic field that actuates the microrobots. Figure~\ref{figure:Experimental_Setup} depicts the experimental setup. While the microrobots used for the experiments have three distinct sections, our simulation simplifies them into single, homogeneous blocks of uniform mass distribution. We argue this assumption is acceptable at the microscale, where factors such as weight and inertia are much smaller in magnitude than factors proportional to distance and surface area, such as adhesion and electrostatic forces. 

\begin{table}[ht]
\caption{Parameters for $\mu$TUM on paper.}
\begin{tabular}{c c c}
\hline \hline
Description   & Value & Units \\
\hline
Mass (m)  & $1.6071\times 10^{-7}$  & kg \\
Electrostatic Force ($F_{elect}$) & $3.2022\times 10^{-6}$  & N\\  
Friction Coefficient ($\mu$) & $0.3$  & -\\
Magnetic Alignment Offset  ($\phi$) & $27$  & degree\\
Magnetic Volume  ($V_m$) & $2.9\times 10^{-11}$  & $m^3$\\
Magnetization ($|\bm{E}|$) & $15000$  & $A/m$\\
Coefficient of adhesion force ($C$) & 3.7148 &$N/m^2$ \\
\hline
\end{tabular}
\label{table_1}
\end{table}

To obtain the adhesion coefficient for the substrate of interest, the microrobot was laid flat over the substrate in dry air. The external magnetic field was set to a static vertical orientation and the field strength was incrementally increased from zero until the microrobot started rotating upwards. The field strength at which rotation occurred was used to calculate the magnetic torque that exactly counteracted the adhesion force resisting upwards motion. Dividing this torque by the moment arm and by the total contact surface area of the robot resulted in the adhesion coefficient for that substrate. To estimate the friction coefficient, a wafer of SU-8 was placed over a sheet of the substrate of interest in dry air. The SU-8 side of the wafer was placed in contact with the substrate and 20 grams of additional mass was attached to the other side, ensuring that the dominant force between the microrobot and the substrate would be weight instead of adhesion or electrostatic forces. The substrate was then tilted from a horizontal position until the wafer started slipping downwards. The angle at which slippage occurred was noted and the friction coefficient for the substrate was  approximated by taking the tangent of this angle.

\begin{figure}[h]
\centering
\includegraphics[width=2.5in]{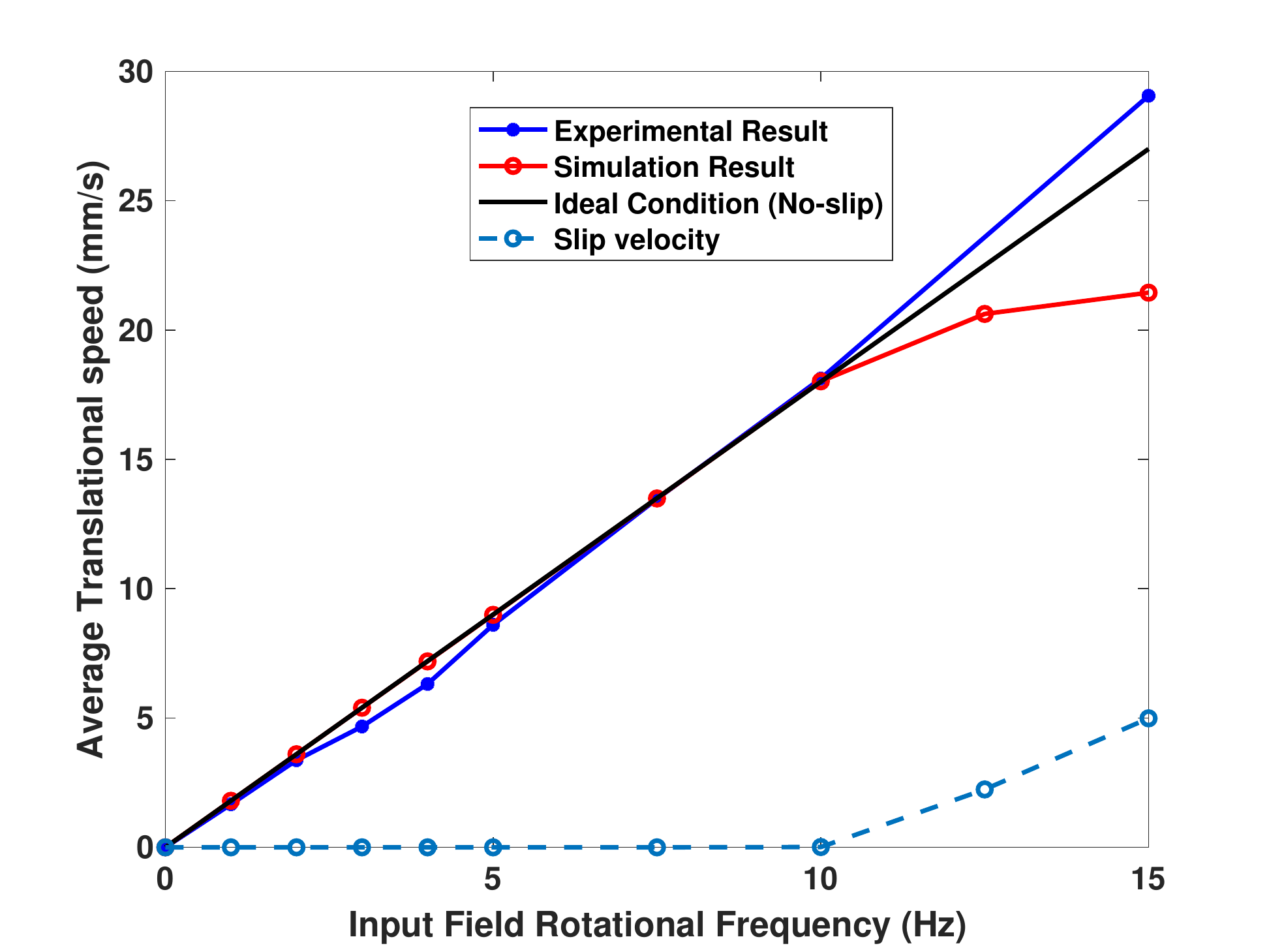}
\caption{Tumbling locomotion tests on paper ($20mT$ field).}
\label{figure:Tumbling} 
\end{figure}

{\bf Tumbling locomotion tests}:
The first scenario investigated is for tumbling locomotion of the $\mu$TUM traversing a dry paper substrate. The parameters are, again, listed in Table~\ref{table_1}. The simulation is performed in order to evaluate the robot's performance on the substrate under varying field rotation frequencies. If the robot tumbles without slipping on the rough paper surface, the robot's average translational speed, $v$, should be approximately equal to two times the sum of body length and body height $(L+H)$ multiplied by the field rotational frequency $f_{rot}$:
\begin{equation}
\label{model_non_slip}
    v = 2(L+H)f_{rot}
\end{equation}
\noindent
In these tests, we compared the state of robot in the experiments and simulation with an ideal no-slip situation. We applied a rotating magnetic field of $20 mT$ to the robot.  Figure~\ref{figure:Tumbling} compares the experimental and simulation results with the ideal no-slip solution (Equation~\eqref{model_non_slip}). When the frequency of rotating field increases, a discrepancy appears between the simulation results and the ideal solution. One possible reason for this discrepancy is that the robot starts slipping on the paper substrate under a high frequency rotational field. We plotted the average slip velocity at different frequencies. As shown in Figure~\ref{figure:Tumbling}, the slip velocity increases as frequency increases, and its value is almost equal to the difference between ideal situation and simulation results. Thus, the discrepancy is mainly due to slip velocity. The experimental results are higher than expected due to complications in the MagnebotiX machine producing the external magnetic field. It is suspected that stray field gradients become more prominent at higher rotational frequencies and pull the microrobot towards the edges of the workspace, causing it to move faster.
\comment{
\begin{table}
\centering
\caption{Inclined plane tests on paper ($20mT$ $@$ $1Hz$).}
\begin{tabular}{c c c}
\hline \hline
Incline ($\theta$)   & Simulation (Y/N) & Experiment (Y/N) \\
\hline
$5^{\circ}$ & Y &   Y    \\
$10^{\circ}$ & Y &   Y    \\
$15^{\circ}$ & Y &   Y    \\
$30^{\circ}$ & Y &   Y    \\
$45^{\circ}$ & Y &   Y    \\
$60^{\circ}$ & N &   N    \\
\hline
\end{tabular}
\label{table_2}
\end{table}
}
\begin{figure}
\centering
\includegraphics[width=2.5in]{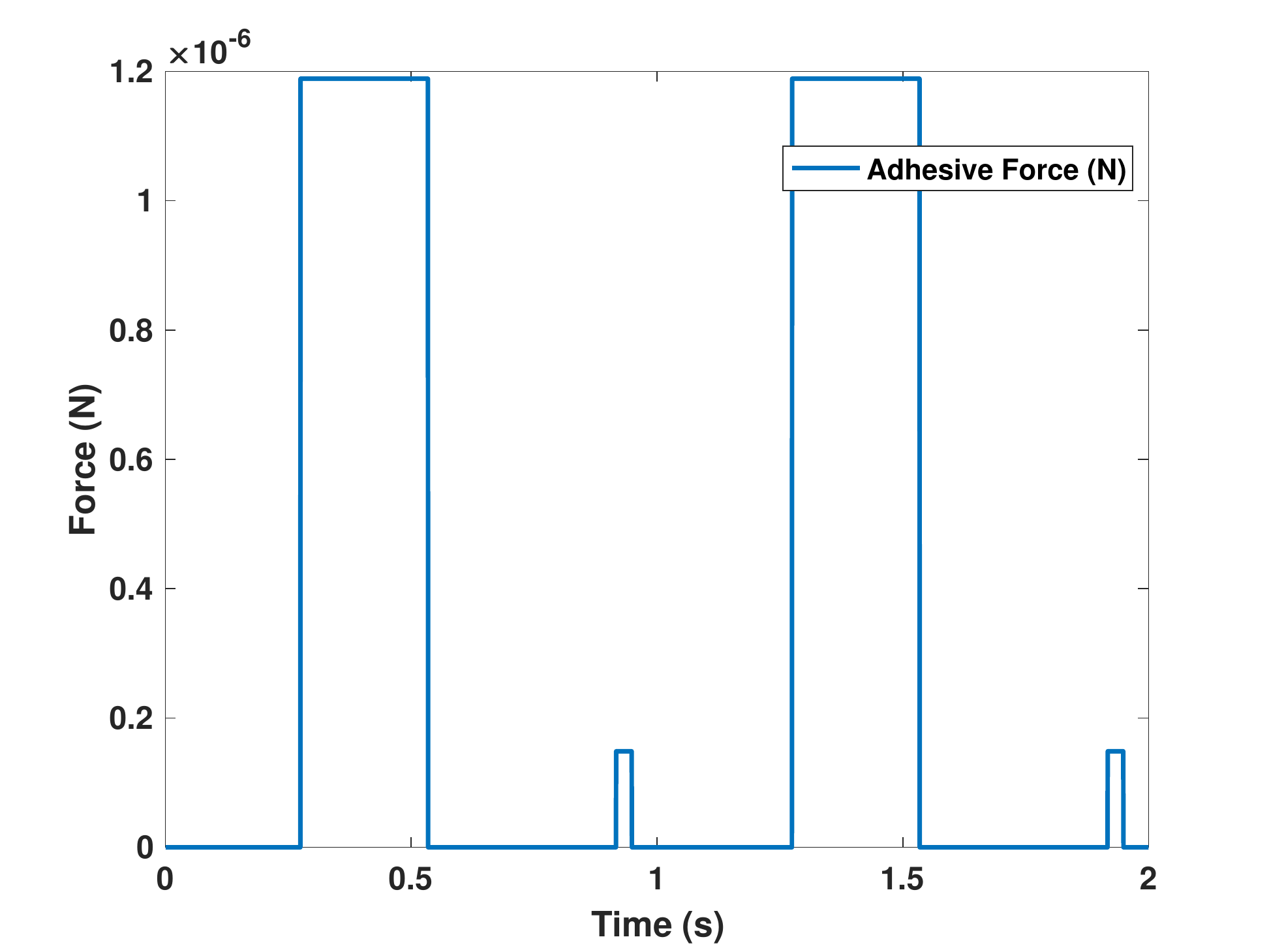}
\caption{Simulation result for adhesive force acting on $\mu$TUM robot when it is tumbling over the incline (paper) of 45 degree ($20mT$ field at $1Hz$). }
\label{figure:ADH} 
\end{figure} 

{\bf Inclined plane traversal tests on paper}:
In the second scenario, the simulation is to determine whether the designed microrobot can climb an inclined surface (paper in dry conditions) at various angles. We applied a $20 mT$ rotating magnetic field and $1Hz$ frequency to the robot. We compared the simulation results with experimental results to validate our model. \comment{The results are reported in Table~\ref{table_2}. }Based on the experimental result, the robots can go over a maximum inclination of $45^{\circ}$ on paper but it will fail to climb a slope of $60^{\circ}$. The simulation output matches these results. Figure~\ref{figure:ADH} plots the adhesive force when robot is tumbling over the incline at $45^{\circ}$. It can be observed from this figure that the force changes periodically. When the contact area is large (Length $\times$ Width), the adhesive force reaches a value of $1.189e-6 N$. When the contact area is small (Width $\times$ Height), the adhesive force value goes to $1.486e-7N$. In line contact cases, the adhesive force is almost zero.

\begin{table}
\caption{Parameters for improved $\mu$TUM on aluminum.}
\begin{tabular}{c c c}
\hline \hline
Description   & Value & Units \\
\hline
Mass (m)  & $6.94\times 10^{-8}$  & kg \\
Electrostatic Force ($F_{elect}$) & $0$  & N\\
Friction Coefficient ($\mu$) & $0.54$  & -\\
Magnetic Alignment Offset ($\phi$) & $0$  & degree\\
Magnetic Volume  ($V_m$) & $3.2\times 10^{-11}$  & $m^3$\\
Magnetization ($|\bm{E}|$) & $51835$  & $A/m$\\
Coefficient of adhesion force ($C$) & 26.1771 &$N/m^2$ \\
\hline
\end{tabular}
\label{table_3}
\end{table}
\comment{
\begin{table}
\centering
\caption{Inclined plane tests on aluminum ($20mT$ $@$ 1Hz).}
\begin{tabular}{c c c}
\hline \hline
Incline ($\theta$)   & Simulation (Y/N) & Experiment (Y/N) \\
\hline
$30^{\circ}$ & Y &   Y    \\
$45^{\circ}$ & N &   N    \\
\hline
\end{tabular}
\label{table_4}
\end{table}
}

{\bf Inclined plane traversal tests on aluminum}:
In our third scenario, we analyze the performance of a $\mu$TUM with improved magnetic properties on aluminum, which is non-magnetic and conductive. Therefore, there shouldn't be any significant electrostatic force or additional magnetic force acting on the robot when it is tumbling over the substrate. Although an electromagnetic drag force may be exerted on the $\mu$TUM due to eddy currents induced in the conductive aluminum, this force is estimated to be two orders of magnitude smaller than the magnetic torque and thus negligible. The coefficient of adhesive force on aluminum was found to be $26.1771 N/m^2$ and the coefficient of friction was found to be $0.54$. The procedure for obtaining the parameters is stated in our experimental setup section. In Table~\ref{table_3}, the magnetization of the newer $\mu$TUM's ($ 51835 \  A/m$) is much higher than that of original $\mu$TUM ($ 15000 \  A/m$). Furthermore, the newer $\mu$TUM has zero magnetic alignment offset angle. We applied a $20mT$ rotating magnetic field at $1Hz$ frequency to the robot. \comment{The result of inclined plane climbing tests are reported in Table~\ref{table_4}.}In both the simulations and the experiments, the robot can successfully climb the inclination of $30^{\circ}$ but fails to climb it at $45^{\circ}$.  A video of showing representative simulation and experimental results can be found here: \url{https://www.youtube.com/watch?v=cr_rrc4NVHE}.

\begin{figure*}%
\centering
\begin{subfigure}[b]{0.66\columnwidth}
\includegraphics[width=\columnwidth]{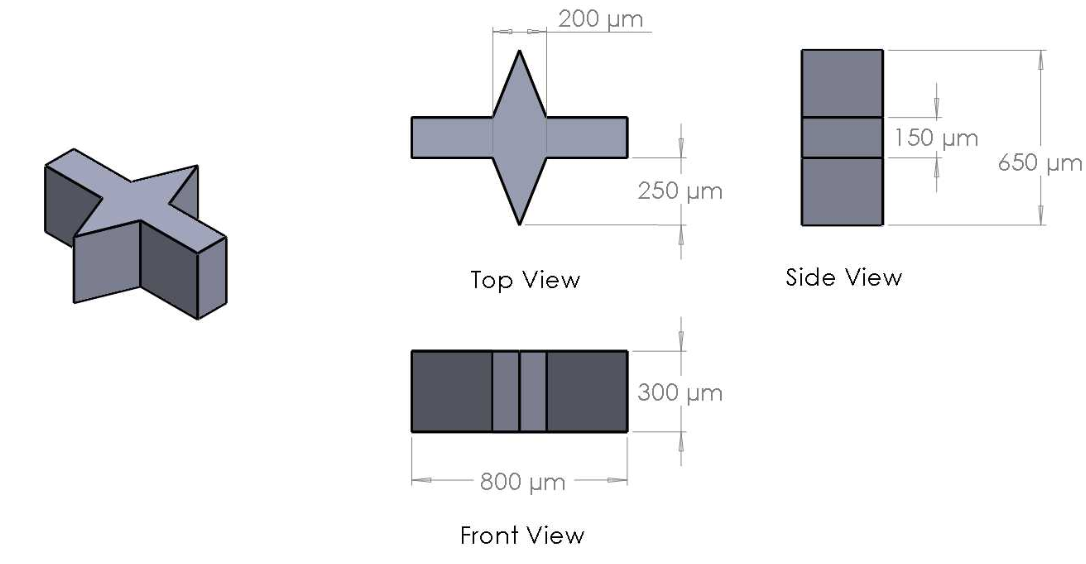}%
\caption{Design for spiked shape $\mu$TUM robot.}
\label{figure_spiked} 
\end{subfigure}\hfill%
\begin{subfigure}[b]{0.66\columnwidth}
\includegraphics[width=\columnwidth]{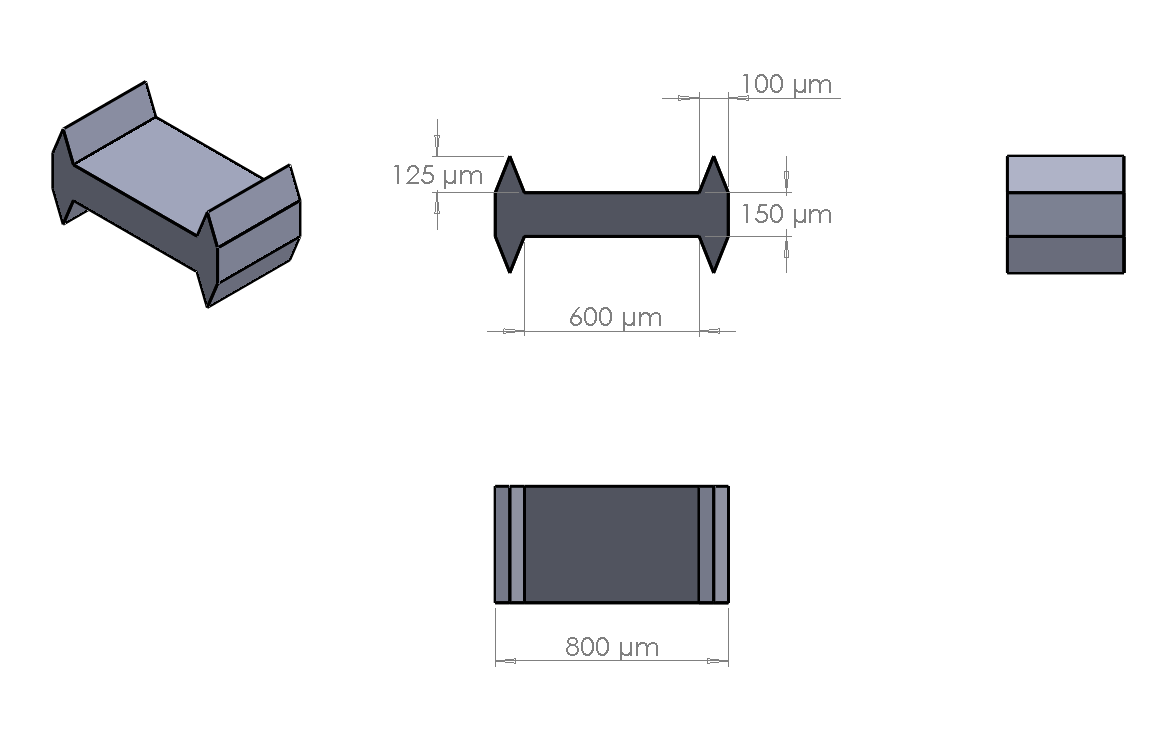}%
\caption{Design for $\mu$TUM robot with spiked ends. }
\label{figure_spiked_ends} 
\end{subfigure}\hfill%
\begin{subfigure}[b]{0.66\columnwidth}
\includegraphics[width=\columnwidth]{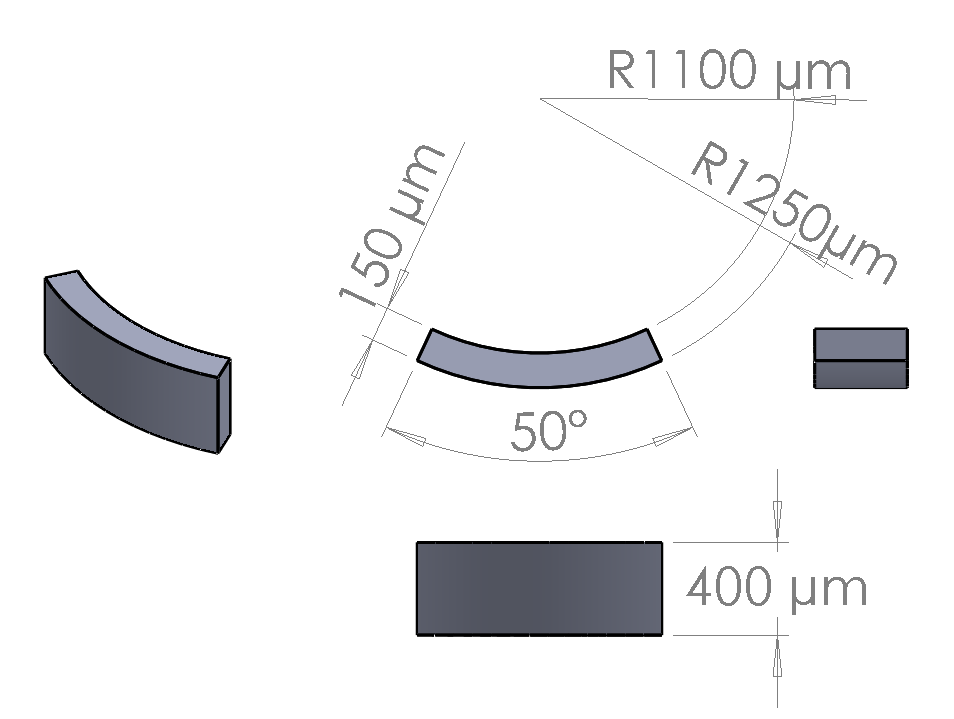}%
\caption{Design for curved shape $\mu$TUM robot. }
\label{figure_curved} 
\end{subfigure}\hfill%
\caption{Design and dimensions of robots with different geometric shapes. }
\label{Dimensions of robots}
\end{figure*}

\comment{
\begin{table}
\centering
\caption{Simulation results for robots with different geometric shapes: inclined plane tests on aluminum ($20mT$ $@$ $1Hz$).}
\begin{tabular}{c c c c c}
\hline \hline
Incline ($\theta$)   & Cuboid  & Spiked  & Spiked Ends  & Curved  \\
 & (Y/N) & (Y/N)& (Y/N)&(Y/N)\\
\hline
$20^{\circ}$ & Y &   Y  &Y  &Y \\
$30^{\circ}$ & Y &   Y  &Y  &N\\
$45^{\circ}$ & N &   N  &N  &N\\
\hline
\end{tabular}
\label{table_5}
\end{table}
}

\begin{figure}%
\centering
\begin{subfigure}[b]{0.5\columnwidth}
\includegraphics[width=\columnwidth]{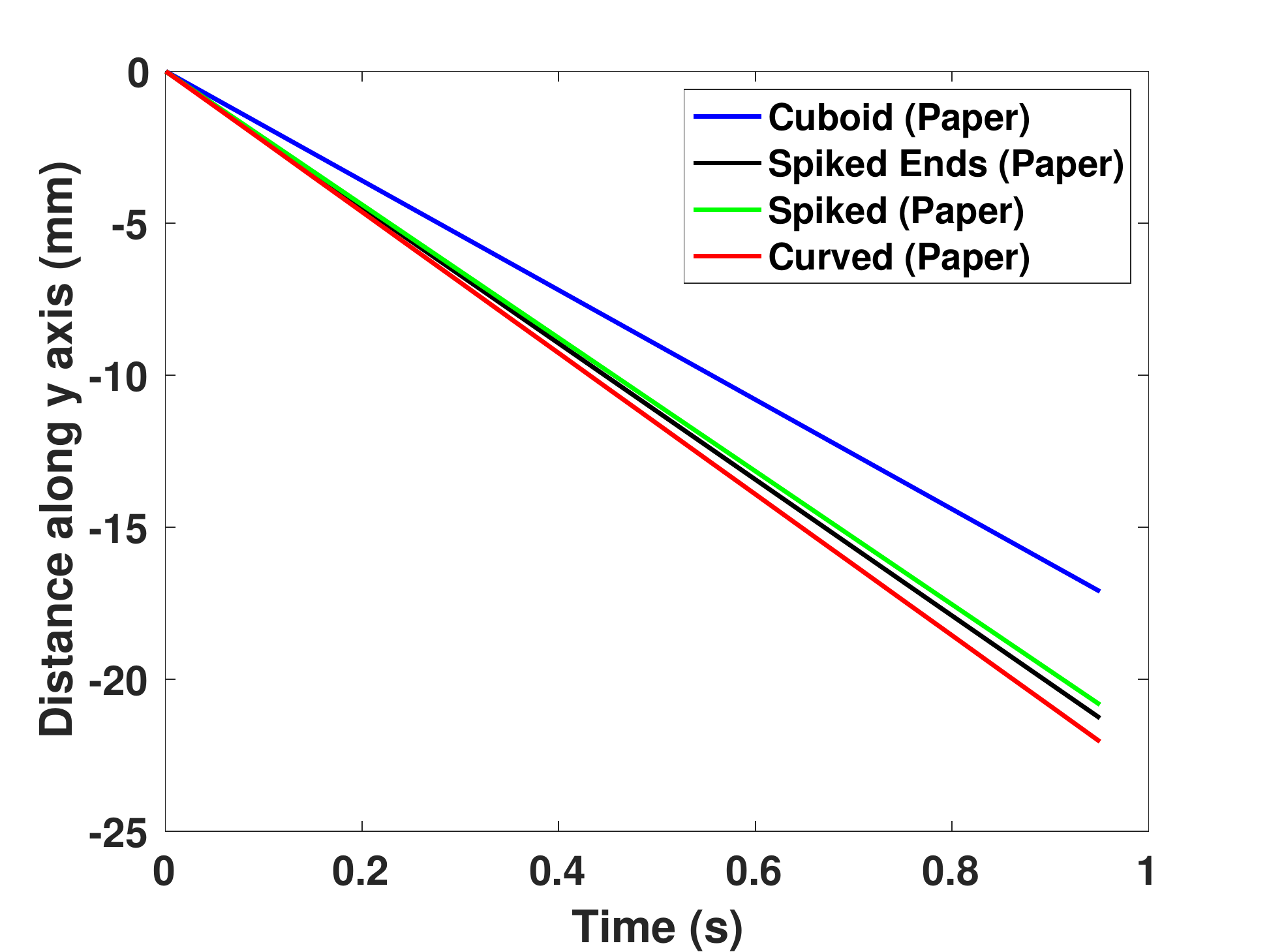}%
\caption{Tumbling tests on paper.}
\label{paper_mul} 
\end{subfigure}\hfill%
\begin{subfigure}[b]{0.5\columnwidth}
\includegraphics[width=\columnwidth]{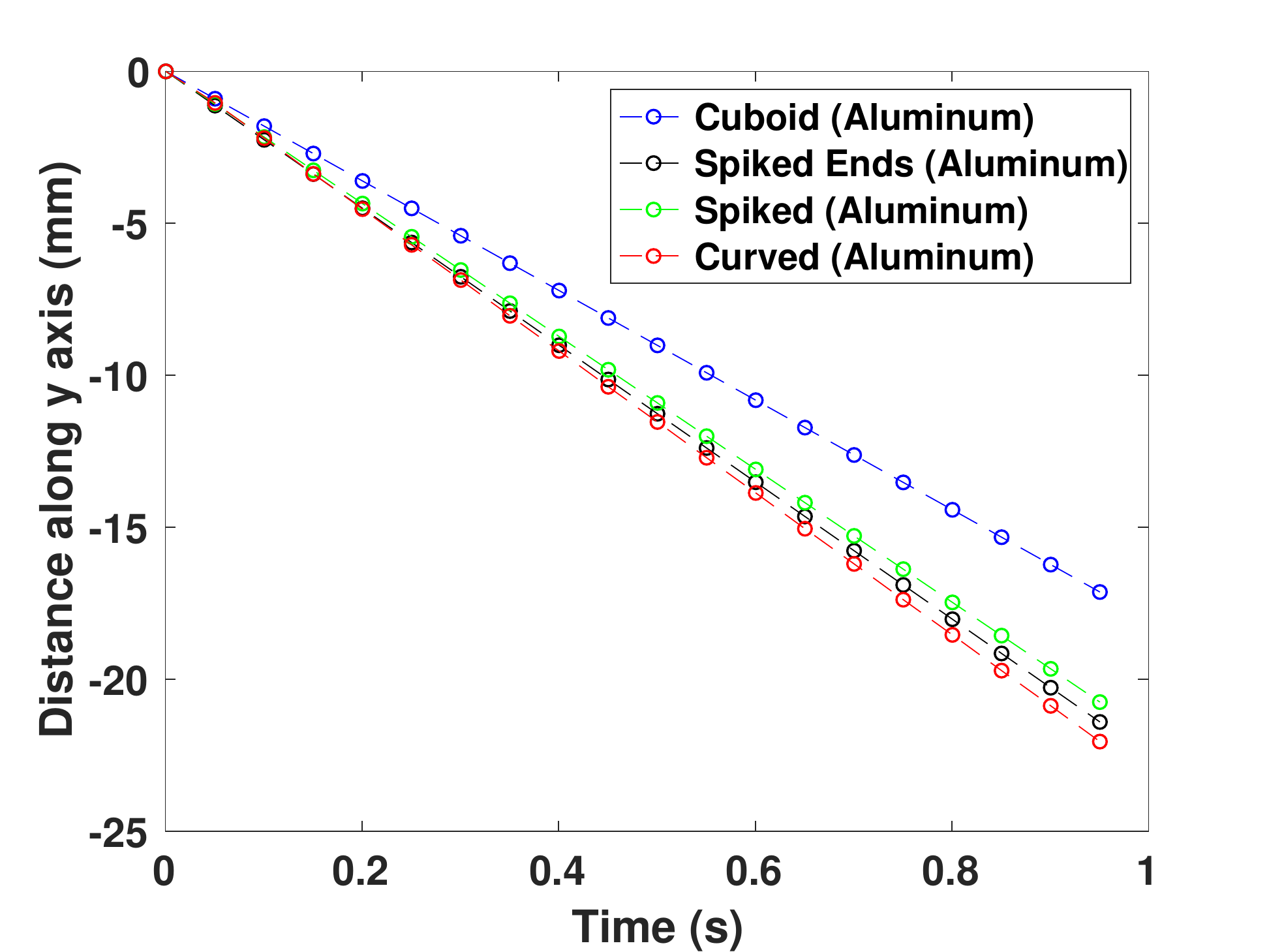}%
\caption{Tumbling tests on aluminum. }
\label{aluminum_mul} 
\end{subfigure}\hfill%
\caption{Simulation result for robots with different geometric shapes: tumbling locomotion test ($20mT$ field at $10Hz$). }
\label{figure:TUMB_MUL}
\end{figure}

{\bf Robots with different geometric shapes}: Now that the simulation model has been validated, it can be used to explore alternative $\mu$TUM geometries for increased performance. As shown in Figure~\ref{Dimensions of robots}, we simulated (a) spiked shape robots, (b) robots with spiked ends, (c) curved shape robots, and (d) cuboid shape $\mu$TUM robot from before. To explore the effect of the robots' design and dimensions on their performance, we assume all the robots all have the same inertia and magnetic properties, the same as those listed in Table~\ref{table_3}.
The simulation includes both the tumbling locomotion tests and the inclined plane traversal tests from before. In the tumbling locomotion tests, we applied a $20 mT$ rotational magnetic field at $10 Hz$ frequency to all the robots. Although these tests could have been performed at $1 Hz$ for consistency, we increased this value to $10 Hz$ in order to emphasize the velocity differences between the four designs due to slip. In Figure~\ref{figure:TUMB_MUL}, each robot's performance on paper is similar to it on aluminum. Furthermore, the curved shape robot was found to move the fastest while the cuboid shape robot moved the slowest. In inclined plane traversal test, we chose the substrate to be aluminum. All robots except the curved shape robot successfully climbed though the incline up to $30^{\circ}$ and fail at $45^{\circ}$. Based on the simulation results, we can conclude that the curved shaped robot performs best in terms of linear speed, but is bad at climbing. In addition, we find that the traditional cuboid shape robot is not the best design for tumbling locomotion. Furthermore, we found that robots with spiked ends geometry has the best overall performance in locomotion tests and inclined plane tests.

\section*{CONCLUSION}
In this paper, we have demonstrated a dynamic simulation model that can account for intermittent non-point contact over multiple substrates and surface inclinations. We validated this model using experiments incorporating a tumbling magnetic microrobot and predicted that spiked ends geometry would result in better overall performance. Using the model as a design aid would help save time and reduce costs on the microrobot iteration and fabrication process. Future developments may include accommodations for soft, elastomeric robot bodies and additional modeling for wet environments.

\bibliographystyle{asmems4}
\bibliography{jiayin}
\end{document}